\documentclass[conference]{IEEEtran}
\usepackage{cite}
\usepackage{graphicx}

\IEEEoverridecommandlockouts
\pubid{\ \hspace*{-9.5cm}Proceedings of IJCNN, July 20-24, Portland, Orgeon, 2003.}

\begin{document}

\title{\Large\bf 
Presynaptic modulation as fast synaptic switching:
state-dependent modulation of task performance
	}

\author{\authorblockN{Gabriele Scheler}
\authorblockA{
ICSI\\
1947 Center Str\\
Berkeley, Ca. 94747\\
Email: scheler@icsi.berkeley.edu}
\and
\authorblockN{Johann Schumann}
\authorblockA{
RIACS/NASA Ames\\
Moffett Field, Ca. 94035\\
Email: schumann@email.arc.nasa.gov}
}

\maketitle

\begin{abstract}

Neuromodulatory receptors in presynaptic position have the ability to 
suppress synaptic transmission 
for seconds to minutes when fully engaged. This effectively alters the 
synaptic strength of a connection. 
Much work on neuromodulation
has rested on the assumption that these effects are 
uniform at every neuron. However, there is considerable evidence 
to suggest that presynaptic regulation may be in effect synapse-specific. 
This would define a second "weight modulation" matrix, which reflects 
presynaptic receptor efficacy at a given site.
Here we explore functional consequences of this hypothesis. By analyzing 
and comparing the weight matrices of networks trained on different aspects 
of a task, we identify the potential for a low complexity "modulation 
matrix", which allows to switch between differently trained subtasks while 
retaining general performance characteristics for the task.
This means 
that a given network can adapt itself to different task demands by regulating 
its release of neuromodulators.
Specifically, we suggest that (a) a network can provide optimized responses 
for related classification tasks without the 
need to train entirely separate networks and (b) a network can 
blend a "memory mode" which aims at reproducing memorized patterns 
and a "novelty mode" which aims to facilitate classification of 
new patterns. 
We relate this work to the known
effects of neuromodulators on brain-state dependent processing. 
 
\end{abstract}

\section{Introduction}
Neuromodulators (NM's) such as dopamine, serotonin or acetylcholine have the 
capacity to activate presynaptic receptors, located at axon boutons and 
involved in the regulation of both glutamate and GABA release 
\cite{PisaniAetal2000}, \cite{UrbanNNetal2002}, \cite{VittenIsaacson2001}.
For the most part, these receptors depress synaptic transmission when they
become activated by a strong neuromodulatory signal.
Neuromodulatory signals are generated by phasic increases of firing of 
e.g. dopamine or serotonin neurons (located in central brain areas such as the 
ventral tegmental area or dorsal raphe) and their effects generally last for 
seconds to minutes \cite{NicolaSMetal2000}, \cite{Aston-Jonesetal2000},
\cite{Rothetal98}.

The plasticity expressed by NM receptors points to a targeted 
regulation at specific presynaptic sites \cite{ScottLetal2002}, 
\cite{TsaovonZastrow2000}. This means that the 
capacity for synaptic depression upon engagement of a NM receptor
will be different for each synapse. The amount of change in 
synaptic strength is governed by the distribution and efficacy of 
presynaptic receptors at a given time. Plasticity in the distribution 
of NM receptors happens on a similar time-scale as long-term potentiation  
(hours for induction, days to weeks at least for retention). Thus 
the distribution of receptors at a presynaptic site is capable of reflecting
experience on a similar time-scale as long-term potentiation, 
which influences the strength of glutamatergic transmission \cite{Scheler2003}.

This paper explores the functional significance of presynaptic 
neuromodulatory receptors and their localization.

We choose conventional, fully trained neural networks as experimental 
material. 
Even though work in computational neuroscience during the past decade has 
shifted the focus within its major paradigm towards the investigation of
precision in spike timing and the importance of  
short-term variability in synaptic transmission, network plasticity is 
still for the most part modelled by long-term potentiation as a way to set 
synaptic weights.

Thus the mechanisms for network plasticity are essentially the same in 
both artificial and biological networks, even though tighter constraints 
on architecture and a limited precision of synaptic weights need to be 
imposed on biological models.

We have therefore opted for conventionally trained neural networks 
as starting points for an investigation on how presynaptic modulation of 
synaptic weights may affect the function of biological networks in a 
state-dependent way.

Behavioral evidence shows that neuromodulators affect performance on 
recognition and learning tasks in ways that are clearly measurable yet 
difficult to conceptualize \cite{BaoSetal2001}, \cite{SpecIssueNN2002}. 
For instance, dopamine and noradrenaline have been linked to 
the ideas of "attention", "arousal", "novelty" and "reward". 
Mathematically, they are usually analysed as regulating a single global 
parameter. This may guide reinforcement learning \cite{SchultzWetal97}, set 
thresholds for signal detection \cite{UsherMetal99} or alter the level of 
(recurrent) connectivity \cite{SchelerFellous2001}. 
Here we propose an alternative mathematical model, the existence of a 
second matrix of stored values designed to be subtracted or added to 
the primary matrix. Obviously, the computational power expressed by weight 
modulation goes considerably beyond that of a global parameter. 
However, we also aim to show that the idea of weight modulation, which 
is synapse-specific and experience-dependent, is entirely 
compatible with the generalized notion of a subtle modulation of task 
execution which leaves basic performance intact. 
Thus we provide a theoretical basis towards further conceptualizing the role of 
neuromodulators in neural processing and the regulation of brain state.

\section{Function of Presynaptic NM receptors}
Synapse-specific modulation of neural transmission means 
that a given weight matrix may be switched to a different but related one
upon engagement of presynaptic receptors. More precisely, this 
"switch" will be often gradual, leading to a blending of stored 
weight values in at least two matrices. (Since there are a number 
of different presynaptic receptors, each targeted by a different substance,
the brain may operate with several modulation matrices. Alternatively, 
we might define a single matrix for the extreme values when all receptors 
are engaged, and the look at the various intermediate states. This 
question is not further addressed in this paper, rather all experiments are 
carried out with a single modulation matrix). 

Since these effects are global on a short time scale throughout the brain,
much work on neuromodulation
has rested on the assumption that these effects are also uniform at 
each neuron or synapse.
But recently, experimental evidence has emerged to the effect that NM 
receptors may indeed be individually regulated by a host of intracellular 
pathways and gene expression mechanisms 
\cite{ScottLetal2002}, \cite{TsaovonZastrow2000}, overlapping with the 
mechanisms that guide glutamatergic strength (such as AMPA receptor regulation).

\begin{figure}[htb]
\centering
\includegraphics[width=7cm,height=3cm]{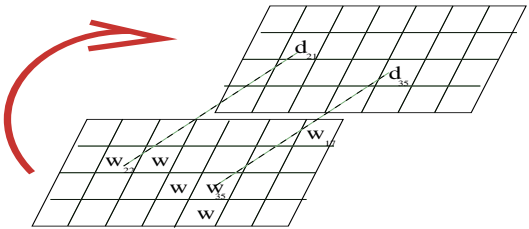}
\caption{Synaptic Switching}
\label{switching}
\end{figure}

The activation of presynaptic NM receptors may be conceptualized 
as fast switching of synaptic weights - where "fast" refers to the time 
required to produce a strong neuromodulatory signal in response to a 
specific stimulus (approximately 100 ms) (Figure~\ref{switching}).

The change will usually last for several seconds to minutes, with 
termination of the effect being 
tightly regulated by a number of complex factors such as re-uptake 
mechanisms, continuing firing of NM neurons and excitation levels of 
the neuron (e.g. calcium and cAMP-levels).

Synaptic switching can be realized by introducing a second matrix that can 
be used to reset specific weights in a primary matrix. Physiologically, 
this corresponds to the capacity of local regulation of NM receptor activation
by both transporter and receptor placement and efficacy. In this way, 
both depression and release from depression of fast glutamatergic/GABAergic 
signalling can be realized by the neural system.

Fast synaptic switching allows specific modulations of task performance 
in trained neural networks.

These modulations become specifically interesting when we are faced with 
a task or aspects of task performance, which cannot be solved by a single 
optimal distribution of weights.

For instance, a set of weights that classifies one set of patterns well may 
be less well adapted for another set of patterns. In this case, rather 
than choosing a weight matrix that covers both patterns in a suboptimal 
way, or learning and maintaining two separate networks for each set of 
patterns, the brain's solution may have been to combine different weight 
sets within a single network, and provide stimulus-specific switching between 
them. Rather than training different networks from scratch, the brain may 
thus reduce training complexity on highly related tasks.
This will work when a basic performance on each aspect of the 
task is guaranteed with either weight distribution. Furthermore, if 
the weights are similar and derived from each other, incomplete switches 
(blends) will produce intermediate results without disrupting
basic task performance.
In a similar vein, the brain's answer to the problem of how to store 
patterns precisely for memorization but also in a more generalized, noisy 
fashion to facilitate classification of novel patterns may have 
been to accommodate both: a set of weights that closely represents a specific 
pattern set, and modifications to this weight matrix to obtain a looser fit,
and promote generalization.
In this paper we present two specific examples for task modification that 
can be realized by synaptic switching between two weight sets each optimized 
for a specific aspect of the task.
The application is taken from the realm of face identification and 
recognition of emotional expressions of faces. The examples presented are 
very simple and designed to exemplify the principle of weight modulation
rather than present a technical solution.
They are primarily meant to illustrate the computational power of presynaptic 
receptors,
once we accept the notion that the localization of NM receptors may be 
functionally regulated, rather than uniformly distributed.

\section{Modulation of Task Performance}
In the first example, we classify a set of patterns  
in a combined task (face identification and recognition of 
emotional expression) by supervised learning. We show that the performance 
of the network for each of the subtasks separately can be improved 
beyond the maximum performance for the whole task. Even though this means
that performance for the other subtask goes down and 
the combined error level remains constant, we have identified a situation,
where basic task performance is guaranteed and synaptic switching allows 
an allocation of precision in memory to one rather than the other 
task. Thus a network can adapt itself towards a focus on face identification 
or a focus on emotional recognition by engaging a neuromodulatory signal 
that subtly alters the weight distribution.

We created a network based on an input representation for 
a face image and trained on both face identification and recognition of 
emotional expression.

The data were taken from a publicly accessible database \cite{faces_DB}.
53 different persons were used, and three different emotions 
(neutral, smiling, crying) were contained in the set (159 different patterns).
Images were scaled and normalized to a size of 20x20 pixels
which comprised the input of the network.
For task A (identification), the output consisted in 6 bits coding for 
53 classes, for task B (emotional recognition), the output consisted in 
3 bits coding for each emotional expression.
A general backpropagation algorithm was used to obtain weight 
matrices for a network with architecture 400x80x9 trained on both tasks 
simultaneously. 
After 4000 iterations of the training set we receive 
a fairly constant result of 
approximately 73\% (face ID) and 86\% (emotional recognition) correctness.
(see Table ~\ref{all-tasks}, combined network).

\begin{table}[htb]
\begin{center}
\begin{tabular}{|l|l|l|l|}
\hline
network &  A & B  & combined\\
\hline
face identification&   93\% & 62\%    & 73\%   \\
emotional recognition& 75\% & 96\%    & 86\%   \\
\hline
\end{tabular}
\end{center}
\caption{Optimization for different aspects of a task}
\label{all-tasks}
\end{table}

The weight matrix is then stored and copied twice. One copy is further 
trained on only subtask A, the other on subtask B. 
This improves performance considerably for either task A or B, and results in 
small losses in the task not trained
(s. Table ~\ref{all-tasks}, network A and network B).

The reduction of training complexity compared to training and storing two 
different networks may not seem significant in the case of a back-propagation
trained neural network. But for a living neural system which takes hours to 
days to change individual synaptic weights, the issue of training time 
(not storage area) is a huge problem. Furthermore the advantage of having a 
coarse, roughly correct system which undergoes subtle modulation as needed
in contrast to a set of highly specialized modules cannot be overestimated.

In a technical sense, whenever we are dealing with a situation, 
where a sequential focus on subtasks occurs, 
but a basic level of performance needs to be maintained 
at all times, this technique of "weight splitting" into two different, but 
similar sets of weights will optimize performance beyond the level of a 
single set of weights and a generalized combined ability.

\begin{figure}[htb]
\centering
\includegraphics[width=7cm,height=5cm]{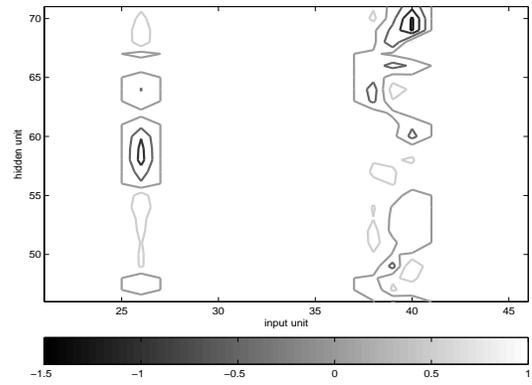}
\caption{Differences between Tasks A and B are localized}
\label{difference-map}
\end{figure}

We may analyze the complexity of the mechanism by a weight difference map for
networks A and B (Figure~\ref{difference-map}).
Very small differences in weight ($<$ 0.03) are not shown.

This results in a picture with significant differences only in certain columns 
rather than others (the figure shows a cutout from the complete network).

\begin{figure}[htb]
\centering
\includegraphics[width=6.5cm,height=5cm]{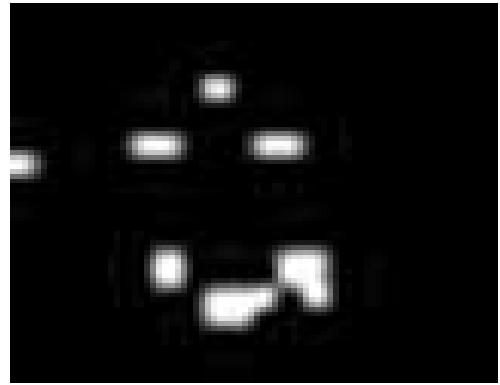}
\caption{Source Units for Strongest Weight Difference between Tasks A and B}
\label{mostdiff}
\end{figure}

We can see that the differences involve selected synapses and 
are fairly local, clustering in certain regions of the input space. 
The source units, connected to these sites of strongest discrepancy are 
are shown in Figure~\ref{mostdiff}, on a 20x20 layout.
In particular, weights encoding features for the eye and mouth region
are affected by changes in the task setting.

This means that the given 
example has a low complexity in the additional training needed for the 
switching mechanism.

We may also compare the hidden representations for selected patterns. 

\begin{figure}[htb]
\centering
\includegraphics[width=4cm,height=2cm]{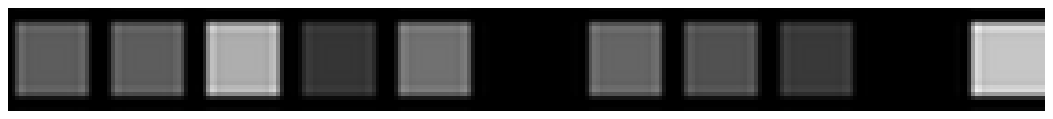}
\includegraphics[width=4cm,height=2cm]{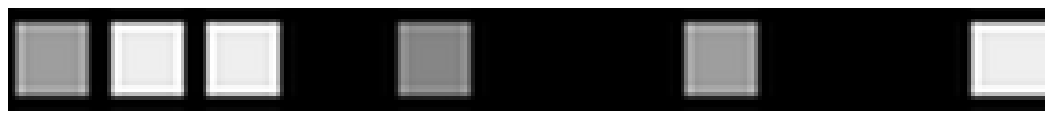}
\includegraphics[width=4cm,height=2cm]{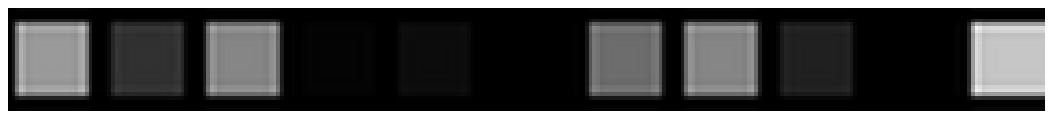}
\includegraphics[width=4cm,height=2cm]{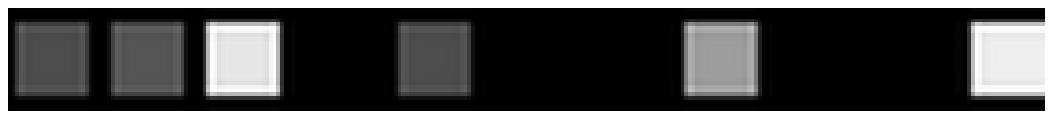}
\caption{Hidden Representations for subtasks and trained networks:
left panel: "smiling" faces , right panel: face no 23,  
upper: network A, lower: network B}
\label{hidden}
\end{figure}

Figure~\ref{hidden}, left panel shows the representations 
that result from superimposing all patterns with "smiling" face expression,
the right panel shows the representations for all patterns for face id \#23. 
In both cases, we see that the representation is similar, but not identical 
for networks A and B. This creates a situation, where blending of two 
networks can be applied without losing basic performance (s. \ref{general}).
In contrast, the effort involved in training two different networks 
independently with essentially highly similar outcomes would not be justified.
Another reason for applying synaptic switching rather than continued training 
of a combinatory task consists in the assumption
that we cannot substitute panel 1, B (optimal) into panel 2, B (suboptimal) 
without affecting panel 2 A (optimal) as well. 
A mathematical analysis of the "restriction of optimality" will help 
to establish this empirical observation.
This should show that certain feature nodes are specifically affected and 
cannot exist in a single "best" position independent of the task that they 
are used for. 

\section{Modulation of Generalization Performance} 
\label{general}

Another modulation that can be implemented with the help of synaptic switching
concerns the trade-off between pattern storage and generalization.
Generally, training a network with optimization for the error level for storing 
a pattern may lead to "overfitting", i.e. a decrease in generalization 
performance, when the learned discriminant becomes too irregular.
A number of techniques have been proposed to influence the 
degree of generalization vs. the storage of patterns (e.g. "early stopping", 
"weight decay" \cite{ReedMarks99}). This trade-off is 
generally regarded to be resolved at the discretion of 
the modeller in accordance with task requirements.

Here we attempt to show that the brain may have implemented this design 
decision with the help of neuromodulation.
The basic idea that neuromodulation may regulate trade-off between 
pattern storage and novel classification has been pioneered by Hasselmo 
\cite{HasselmoBarkai95}, where the self-organization of feedforward 
connections was described as benefiting from suppression of strongly 
modified intrinsic connections associated with specific prior learning.
The mechanism proposed here is more general, but a state-dependent 
modulation of learning vs. storage optimization seems to be one of the tasks 
of neuromodulation.

We select a training and a test set from the face identification problem.
The training set consists of 100 patterns (2 for each face) and the test set 
of 50 patterns (1 for each face randomly selected).

Using a weight-decay backpropagation algorithm, we first obtain
a network which performs well on the training set and minimizes the 
error in generalization (see Table \ref{overfitting}, network A,
1500 iterations, architecture of the network is 400-10-3).

\begin{table}[htb]
\begin{center}
\begin{tabular}{|l||l|l|l|l|}
\hline
network & training & generalization  \\
\hline
A (trained for generalization)   & 71\% & 72\%\\
B (trained for memorization) & 98\% & 68\%\\
\hline
\end{tabular}
\end{center}
\caption{Overfitting: \% of correct patterns for more or less highly trained 
networks}
\label{overfitting}
\end{table}

Then, we perform additional training (without weight decay) to improve 
the network's capability to recall the training data (network B, 
additional 4000 iterations). 
This training 
results in 98\% correct identification of the faces in the training set, 
but slightly decreases the generalization performance.
By this method, as in the previous example, we obtain two different networks
that are highly similar but different enough in selected synaptic weights 
to subtly alter task performance.

\begin{figure}[htb]
\centering
\includegraphics[width=7cm,height=5cm]{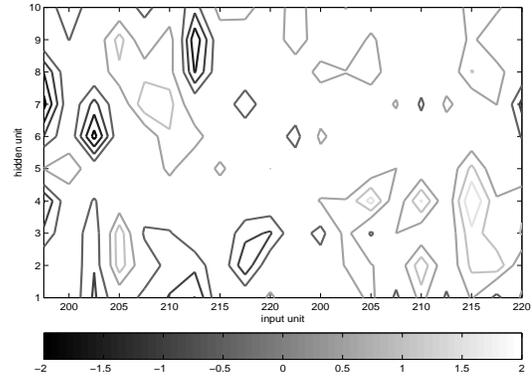}
\caption{Weight difference diagram for networks A and B.}
\label{gendd}
\end{figure}

The comparison of the difference in weights between the two networks is shown
in Figure~\ref{gendd}. There is no clearly discernible structure to the 
weight difference diagram, thus we would expect complexity to be higher 
in this case.

An interesting possibility that is supported by the physiological evidence 
is partial weight modulation. There are essentially two different mechanisms 
for that, some combination of which probably occurs in the brain. One 
mechanism assumes a partial activation of receptor sites by a limited increase 
of neuromodulator availability. This would result in a linear change of 
weight values. The other mechanism assumes that only a percentage of receptor 
sites are activated fully - other receptors being decoupled or desensitized.
This would result in a potentially skewed change in weight values.

\begin{figure}[htb]
\centering
\includegraphics[width=7cm,height=5cm]{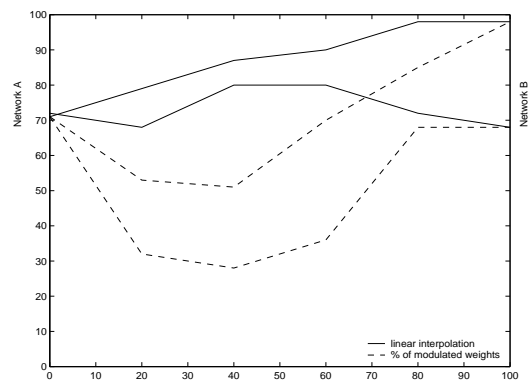}
\caption{Partial Weight Modulation: Effects on performance}
\label{blending}
\end{figure}

Figure~\ref{blending} shows the effects of both techniques on performance 
measures for the generalization-storage trade-off. Endpoints of the 
trajectories for storage and generalization are given by the values in 
Table~\ref{overfitting}. Interestingly, shutting off a percentage of 
the receptors leads to fluctuations in performance (dashed line), while the 
linear interpolation approximates a corresponding linear change in 
performance (continuous line).

\begin{figure}[htb]
\centering
\includegraphics[width=4cm,height=3cm]{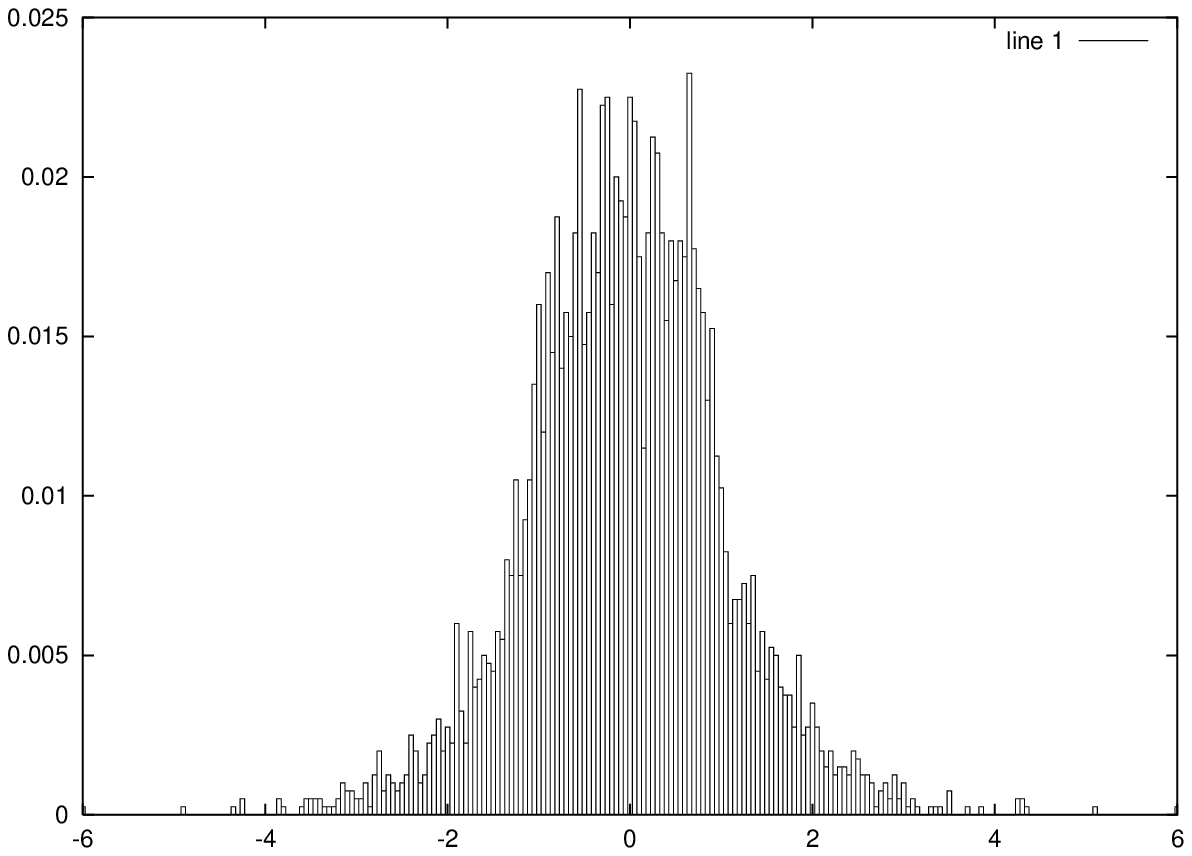}
\includegraphics[width=4cm,height=3cm]{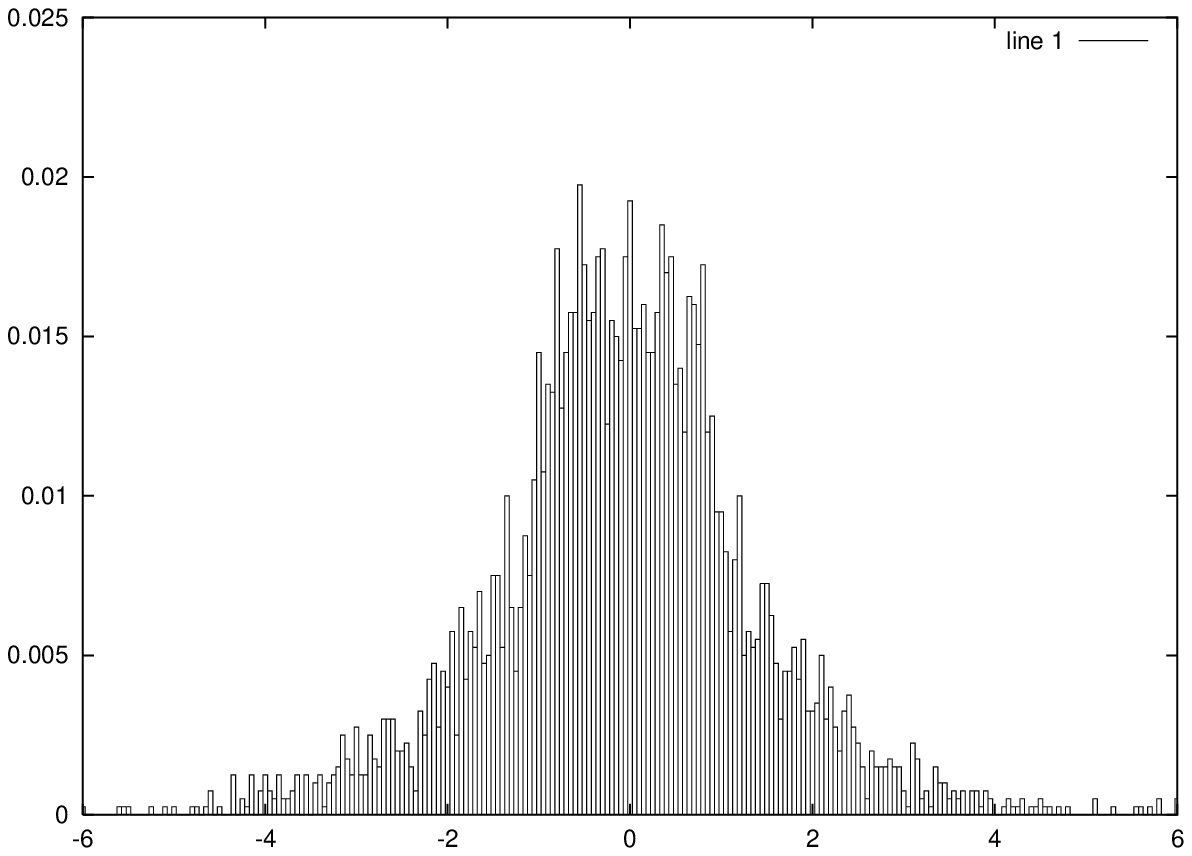}
\caption{Weight distribution for networks A (left) and B (right).}
\label{weight-distr}
\end{figure}

Finally, another way to compare the weight matrices that result 
from training for storage versus training for generalization is to look
at the distribution of actual weight values (Figure~\ref{weight-distr}). 

We can see that the variance for the generalizing network is lower (1.2) 
than for the storing network (2.0). This is in accordance with the 
observation that weight decay helps in achieving better generalization.
Physiologically, weight modulation may have the side effect of 
decreasing the range of synaptic strengths. Here we can see that this 
feature may have been applied in a functionally useful way by the brain.

\begin{figure}[htb]
\centering
\includegraphics[width=4cm,height=3cm]{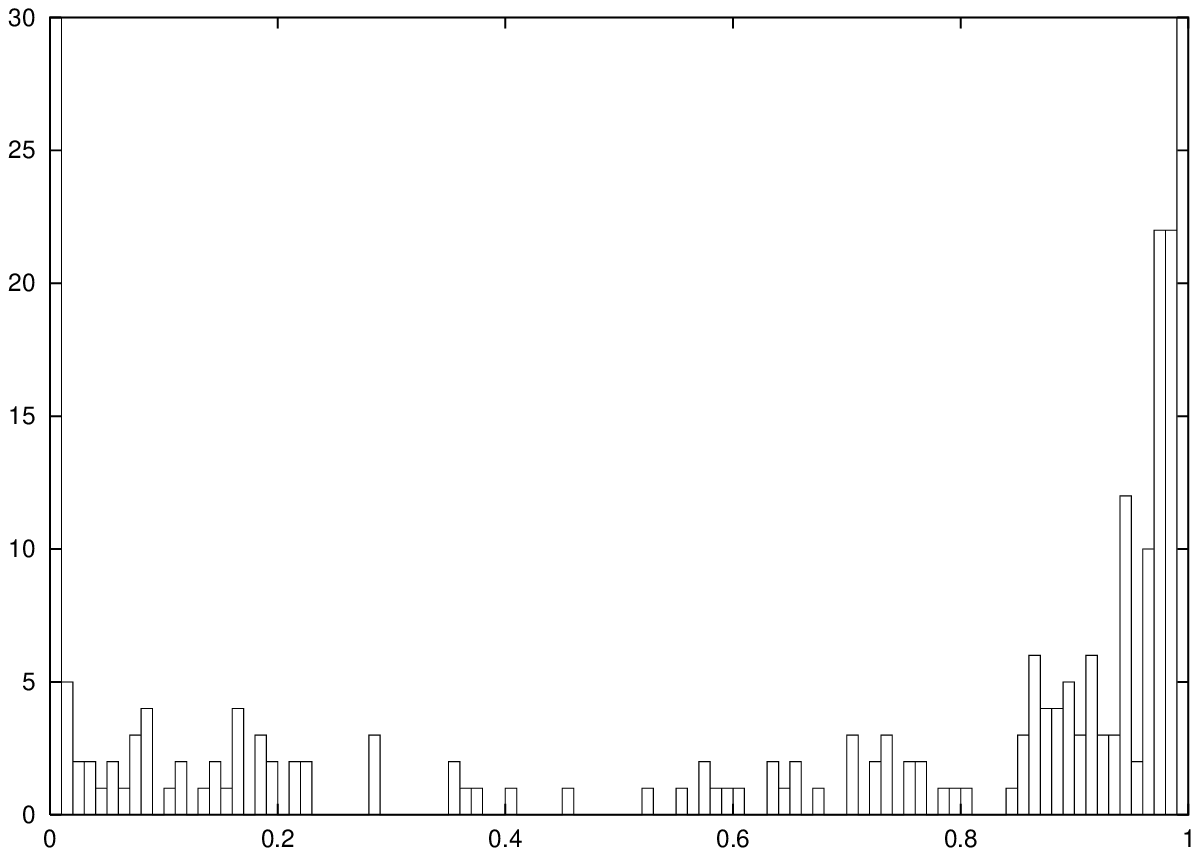}
\includegraphics[width=4cm,height=3cm]{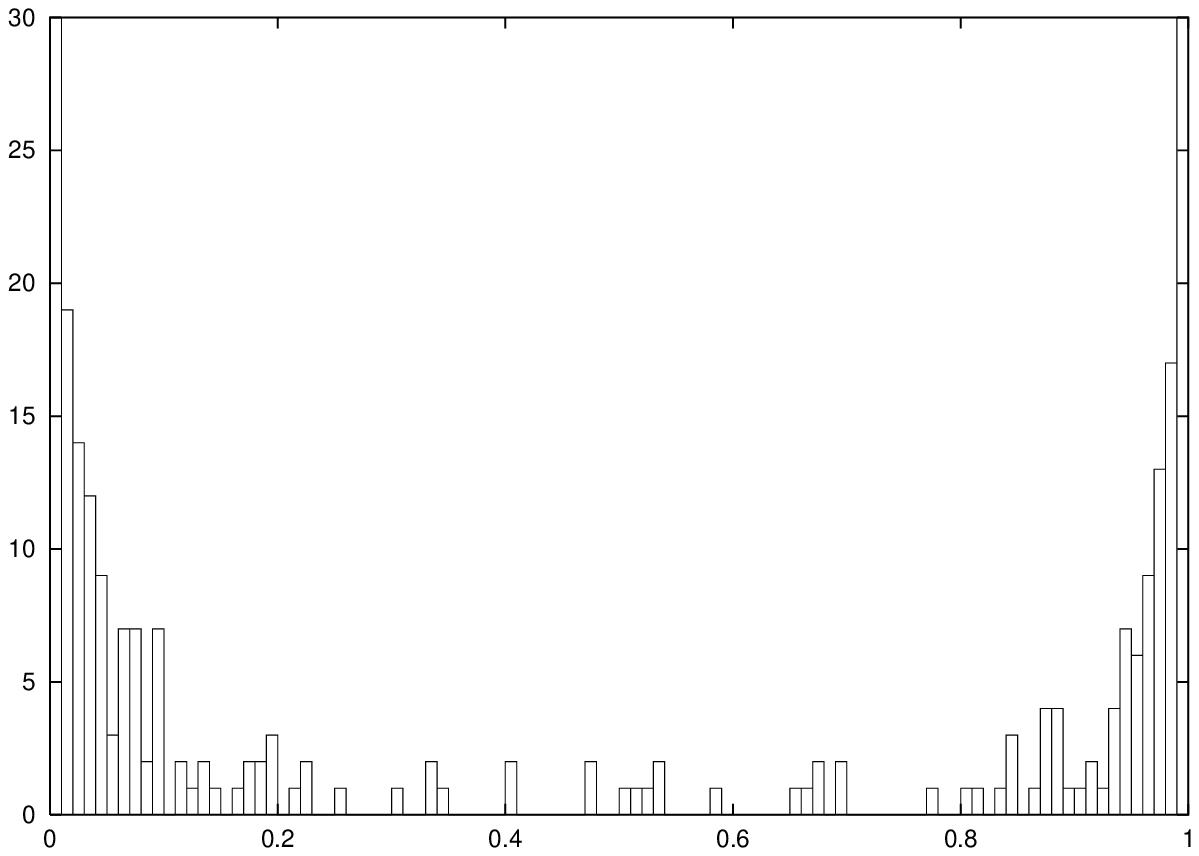}
\caption{Firing rate distribution for networks A (left) and B (right).}
\label{ratedist}
\end{figure}

Similarly, we may compare the firing rate distribution for both networks
(Figure~\ref{ratedist}). 

This shows that the network optimized for storage has more 
neurons with low activation than the network which has been optimized for 
generalization. Again, this may point to a focusing of activation on 
selected neurons and less distributed activity in the network, which is 
a result that is compatible with neuromodulatory alterations of 
network activity.

\section{Conclusion}
In general, neuromodulators define brain state and alter neural processing 
according to current needs of the organism. 
Often it is assumed that the effects of NM are uniform 
at all synapses. In this sense, the modulation mediated by e.g. dopamine 
receptors depends only on a global signal, namely phasic increase of dopamine 
neuron firing and release. Accordingly, the function of neuromodulation
has been linked to general, unspecific alterations in processing mode, 
such as increased signal-to-noise ratio, 
vigilance or arousal \cite{Koch99}, (p.225-226),
a general 
reinforcement signal \cite{AbbottDayan2001}, (p. 339-340) 
or increased recurrent connectivity \cite{SchelerFellous2001}, 
\cite{DurstewitzSeamans2002}.

Here we show that the experimental evidence which supports localized responses 
greatly enhances the computational power associated with neuromodulation.

In particular, the technique of fast synaptic switching to a second weight 
matrix can be applied to increase performance levels of related tasks 
individually. We have applied this to classification of faces according to 
identity versus recognition of an emotional expression and to the memorization
of face images versus the ability to classify novel images.

The basic idea of fast synaptic switching is not novel. A related form of 
synaptic switching 
within a neural processing network has been explored in the context 
of the "dynamic link architecture" \cite{BienenstockDoursat94}, 
\cite{Konenetal94}, \cite{AonishiKurata98}. 
The dynamic link architecture has been mostly used for the extraction and 
storage of invariants in perceptual processing. Its possible link to 
the physiological substrate of neuromodulation has not been explicitly explored.
But the dynamic link architecture incorporates techniques for learning 
not only the primary weight matrix, but also a secondary matrix which stores 
information on the target weights that undergo switching.
Our work has not addressed the question of a "learning rule" for the weight modulation 
matrix, i.e. the placement of presynaptic receptors. Rather we have explicitly 
constructed complete, fully trained weight matrices by conventional means, 
and explored the consequences of being able to blend or switch them by 
neuromodulatory signals. We have however made the observation that the 
complexity of learning can be expressed by the number of receptors that have 
to be placed.

We have taken care to ensure that the results are compatible with the 
forms of state-dependent processing which have been documented as behavioral 
modifications due to neuromodulatory function.
Subtle alterations in task performance due to 
engagement of neuromodulator receptors provide a form of adaptivity 
that ensures basic performance but 
allows task-specific optimization. We feel that this description of 
neuromodulatory function provides a framework for further experimental 
and theoretical studies.

{\bf Acknowledgments:}
We want to thank Pramod Gupta, Tony Bell, and two anonymous reviewers 
for helpful suggestions during the preparation of this manuscript. 
This work was supported in part by project NASA ECS RSO "Adaptive 
Control Technologies".

\end{document}